\documentclass[lettersize,journal]{IEEEtran}
\usepackage{amsmath,amsfonts}
\usepackage{algorithmic}
\usepackage{array}
\usepackage[caption=false,font=normalsize,labelfont=sf,textfont=sf]{subfig}
\usepackage{textcomp}
\usepackage{stfloats}
\usepackage{url}
\usepackage{verbatim}
\usepackage{graphicx}
\def\BibTeX{{\rm B\kern-.05em{\sc i\kern-.025em b}\kern-.08em
    T\kern-.1667em\lower.7ex\hbox{E}\kern-.125emX}}
\usepackage{balance}

\usepackage{latexsym}

\usepackage[T1]{fontenc}

\usepackage[utf8]{inputenc}

\usepackage{microtype}

\usepackage{setspace}

\usepackage{svg}

\usepackage{tabularray}

\usepackage{xstring}

\usepackage{graphicx}
\usepackage{subcaption}

\usepackage{caption}

\usepackage{tabularx}
\usepackage{tabulary}

\usepackage{multirow}
\usepackage{color}
\usepackage{booktabs}
\usepackage{multirow}
\usepackage{enumitem}
\usepackage{stfloats}
\usepackage{pifont}

\usepackage[most]{tcolorbox}
\newtcolorbox{prompt_box}[2][]{%
  attach boxed title to top left
               = {yshift=-1.5pt},
  colback      = black!5!white,
  colframe     = black!75!black,
  colbacktitle = gray!85!black,
  title        = #2,#1,
  enhanced,
  boxsep=1pt, 
  left=1mm,   
  right=0mm,  
  top=1mm,    
  bottom=1mm  
}
\usepackage{color,soul}
\usepackage{eso-pic}

\begin{document}

\AddToShipoutPictureFG*{
\AtPageUpperLeft{
\put(45,-27){
\parbox{0.9\textwidth}{
\centering
\fbox{
\parbox{0.95\textwidth}{
\footnotesize
© 2026 IEEE. Personal use of this material is permitted.  Permission from IEEE must be obtained for all other uses, in any current or future media, including reprinting/republishing this material for advertising or promotional purposes, creating new collective works, for resale or redistribution to servers or lists, or reuse of any copyrighted component of this work in other works.

This is a preprint of the accepted paper for publication in IEEE Transactions on Computational Social Systems.
}
}
}
}
}
}

\title{Debiasing Large Language Models toward Social Factors in Online Behavior Analytics through Prompt Knowledge Tuning}

\author{Hossein Salemi, Jitin Krishnan, Hemant Purohit
\\ George Mason University, Fairfax, Virginia, United States
}

\maketitle

\begin{abstract}
Attribution theory explains how individuals interpret and attribute others' behavior in a social context 
by employing personal (dispositional) and impersonal (situational) causality. Large Language Models (LLMs), trained on human-generated corpora, may implicitly mimic this social attribution process in social contexts. However, the extent to which LLMs utilize these causal attributions in their reasoning remains underexplored. 
Although using reasoning paradigms, such as Chain-of-Thought (CoT), has shown promising
results in various tasks, ignoring social attribution in reasoning could lead to biased responses by LLMs in social contexts. In this study, we 
investigate the impact of incorporating a user's goal as knowledge to infer dispositional causality and message context to infer situational causality on LLM performance. 
To this end, we introduce a scalable method to mitigate such biases by enriching the instruction prompts for LLMs with two prompt aids using social-attribution knowledge, based on the \textit{context} and \textit{goal} of a social media message. 
This method improves the model performance while  
reducing the social-attribution bias of the LLM in the reasoning on zero-shot classification tasks 
for behavior analytics applications. 
We empirically show the benefits of our method across  
two tasks--intent detection and theme detection on social media in the disaster domain--when considering the variability of disaster types and multiple languages of social media. 
Our experiments highlight the biases of three open-source LLMs: \textit{Llama3}, \textit{Mistral}, and \textit{Gemma}, toward social attribution, and show the effectiveness of our mitigation strategies. Source code and datasets will be shared upon final acceptance of the paper. 
\end{abstract}


\section{Introduction}
\IEEEPARstart{R}{etrieving} intrinsic knowledge embedded in Large Language models (LLMs) to use in Natural Language Processing (NLP) 
problems 
is  a critical component in engineering efficient computational social systems using Artificial Intelligence (AI) \cite{10.1145/3560815}. Especially, when LLMs are utilized for designing real-world  
applications in zero-shot settings, where there is 
no data for training the model, background knowledge plays an important role. Recent works have shown that instruction-following LLMs could leverage 
their background knowledge to improve zero-shot performance in tasks such as sentiment analysis \cite{zhang-etal-2024-sentiment}
and stance detection \cite{doi:10.1073/pnas.2305016120}. Zero-shot Chain-of-Thought \cite{kojima2023large} (referred to as Zero-CoT in this paper) is one of the state-of-the-art reasoning methods that leverages the chain-of-thought capability of LLMs in a zero-shot setting to effectively utilize LLMs' background knowledge and improve the performance of the LLMs' inference. 
However, LLMs have been shown  
to have some intrinsic biases in the training phase and 
biases in prompts for instructing a LLM in the inference phase \cite{10.1145/3539597.3570382}. Thus, instruction-following LLMs require proper contextual prompts that 
help these models retrieve relevant knowledge to aid their inference process.

\begin{figure*}[tp]
  \centering

\includegraphics[width=\textwidth]{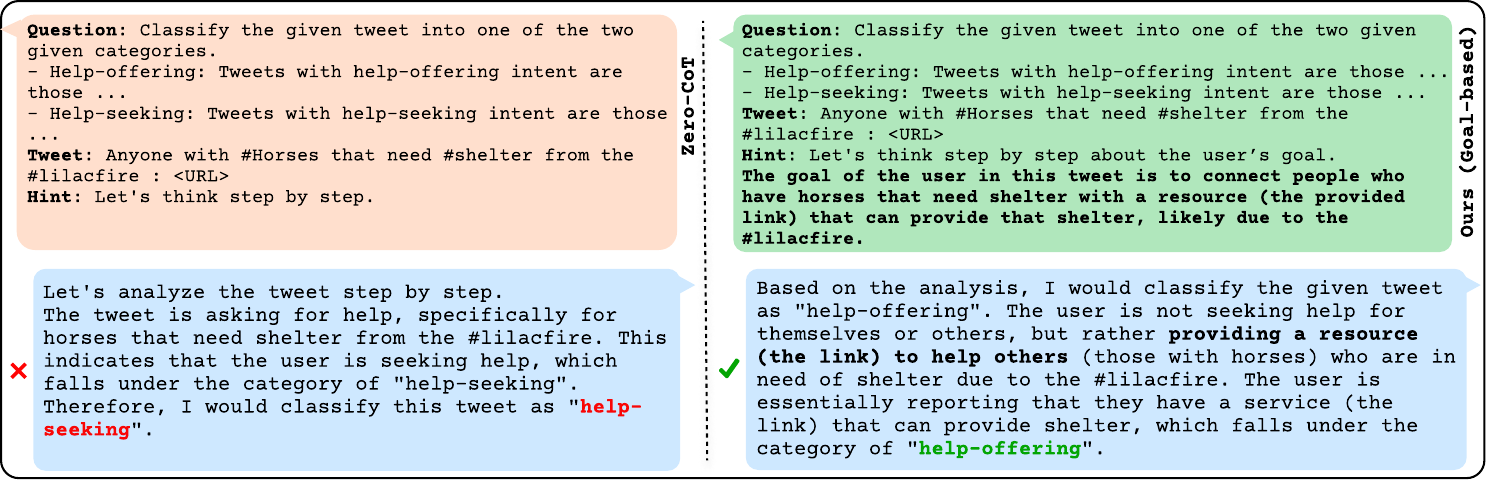}
\caption{\label{fig:main-dialogue}Example of our Goal-based prompt for intent detection task to alleviate the social-attribution bias of Llama3 model by providing the knowledge aid of social factors, i.e., the user's goal in this prompt.
}
\end{figure*}

Moreover, using instruction-following LLMs can be more challenging when they are used in 
ambiguous language understanding tasks such as  
behavior and affect analyses from content~\cite{zhang-etal-2024-sentiment}. This complexity can originate from the social context, a specific domain of the task, or the innate complexity of the given task. For instance, in the disaster domain, for predicting the intent of a 
social media
post  
in which the user shares a link (like Fig. \ref{fig:main-dialogue}), the LLM should consider the post  
from two perspectives to infer \textit{help-offering} class as the intent: 1) what the user is doing (sharing information), 2) the cause of this behavior in the disaster situation (helping via informing people during disaster). While the former can be realized through analyzing the content 
by LLMs, understanding the latter requires paying attention to the social aspects--
the user's goal in this case. The way that individuals interpret and attribute the cause of a behavior can be explained based on the \textit{Attribution Theory} \cite{heider_psychology_1958} rooted in social science. Based on this theory, people may attribute the cause of a behavior to the characteristics of the person (\textit{personal or dispositional causality}), such as the person's desire and intent, or to the environmental events that cause the behavior (\textit{impersonal or situational causality}). Since 
LLM agents are trained on human-generated corpora also, they are likely to replicate such social attributions in their reasoning. For instance, in Fig. \ref{fig:main-dialogue}, given our prompt strategy, the model attributes the behavior of sharing the link to the intention of the user to help others, which mimics the dispositional causality by humans, while the Zero-CoT strategy fails to capture 
this causality. This example shows that while 
a  
model can incorporate 
such   
causal reasoning to improve its outputs, it requires knowledge aids (user's goal in this case)  
for 
accurate
reasoning. 
Previous NLP studies have reported the neglect of social attributions, 
as \cite{hovy-yang-2021-importance} states: 

\begin{quote}
    ``... current NLP systems still largely ignore the social aspect of language. Instead, they only pay attention to what is said, not to who says it, in what context, and for which goals.'' 
\end{quote} 

\begin{figure}[t]
    \centering
    \includegraphics[width=0.9\linewidth]{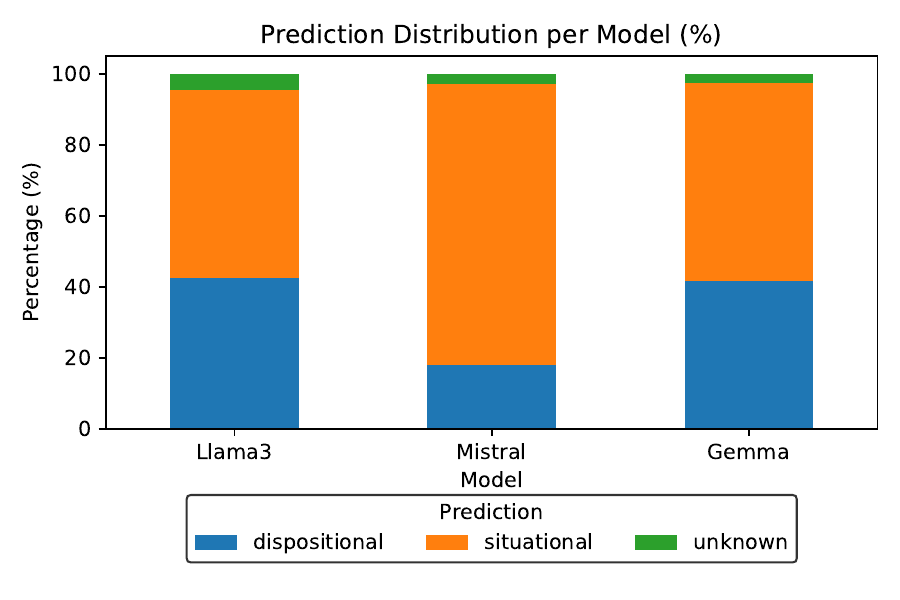}
    \caption{Distribution of significant social attribution causalities in the \textit{Intent} task across models, highlighting the underestimation of dispositional causality.}
    \label{fig:bias_justification}
\end{figure}

Therefore, one limitation of language models is that they sometimes focus more on lexical cues 
while disregarding critical social factors such as context and the user's goal; this imbalance 
can adversely affect the performance of LLM-based AI systems in zero-shot classification tasks. Accordingly, in this paper, we define \textit{social-attribution bias} as the tendency of LLMs to rely unevenly on dispositional (e.g., user goals) and situational (e.g., contextual) causal explanations in social reasoning.  
Notably, appropriate attribution depends on the task: intent detection primarily requires dispositional causality, whereas theme detection relies more heavily on situational causality; social-attribution bias arises when LLMs fail to adjust this causal focus accordingly. 
For motivation,  
we conducted an experiment in which, given the task description, we prompted LLMs to indicate which type of causality is more salient for the task. Fig.~\ref{fig:bias_justification} presents the results
\footnote{We used 10 semantically equivalent prompts and generated 20 responses per prompt using sampling-based decoding (temperature $T=0.7$, top-$p=0.9$), yielding a total of 200 responses per model.} for the intent detection task across three LLMs. We observe that all models underestimate the significance of dispositional causality for intent detection task, highlighting their social-attribution bias.

In this work, we investigate the social-attribution bias of open-source instruction-following LLMs and show how these biases can reduce the performance of LLMs on our behavioral analytics tasks. 
We argue that enriching the prompt with suitable knowledge aids derived from the social aspects (text's context and user's goal) of the textual message is an effective method to mitigate the social-attribution bias and enhance the performance of LLM-based AI systems in such tasks. 
We hypothesize that providing the user's goal as the knowledge aid that represents the user's desire helps the model to leverage dispositional causality, 
while
providing the message context, which describes the environmental events 
in the messaging 
enables the model to rely on situational causality in its reasoning.

Providing these knowledge aids 
presents
another challenge that needs to be addressed.  
Since manually 
generating these knowledge aids is not practically feasible 
at scale, 
we propose a simple 
yet 
effective solution. 
We hypothesize that a 
LLM may 
already
contain some of this knowledge, 
however,
the LLM 
is unable to  
leverage 
it
during the inference stage. Therefore, 
retrieving this knowledge from the LLM itself and providing it explicitly 
through the prompt improves the LLM's performance in the zero-shot setting.  
In this paper, we 
address the following research questions to evaluate the performance of LLM-based AI systems in domain-specific tasks in social computing:

\textbf{RQ1:} Are LLMs biased toward leveraging social factors (context and users' goal) that impacts their reasoning and degrades their performance in zero-shot domain-specific tasks?

\textbf{RQ2:} Can additional prompt aids based on social-attribution knowledge of the LLM to accurately interpret a text helps the LLM to mitigate the social-attribution biases and improve their performance in zero-shot domain-specific tasks?

\textbf{RQ3:} Can the proposed knowledge-based prompt strategies improve an LLM's performance in multi-lingual settings? 
 
\textit{\textbf{Contributions.}}
        Our main contributions are as follows:
        \begin{itemize}
            \item We analyze the social-attribution biases of three instruction-following large language  
            models ($Llama3$, $Mistral$, and $Gemma$) in two zero-shot domain-specific 
        tasks, 
            including
           intent detection and theme detection on social media messages in disaster domain.

           \item We extend the Zero-CoT pipeline with a pre-reasoning knowledge extraction step that automatically retrieves the LLM’s social-attribution knowledge about the message. This scalable approach enables the model to incorporate 
           the retrieved 
           knowledge during reasoning and mitigate social-attribution bias. We compare the performance of our mitigation strategy and the Zero-CoT strategy to highlight the LLMs' biases toward these social attributions. 
           
           \item To this end, we design templates for prompting LLMs with diverse prompt aids based on social-attribution knowledge of \textit{context} and \textit{goal} of the message in the zero-shot setting.

           \item We evaluate our strategies through ablation studies to assess the impact of key factors on prompt strategies in mono-lingual and multi-lingual settings of disaster domain, as the case study. The results provide insights into the potential of our method in designing more effective LLM-based 
           computational social systems for disaster 
           domain. 
        \end{itemize}

In the rest of the paper, we first review related work, and explain our methodology and the proposed prompt aid strategies based on social-attribution knowledge. We then describe the datasets and experimental setup. Finally, we 
discuss
the results and present the findings of ablation studies.

\section{Related Works}

\begin{table}[t]
\centering
\footnotesize
\caption{Comparison of our method with prior context injection and external knowledge prompting approaches.}
\begin{tabular}{p{0.04\textwidth}p{0.14\textwidth}p{0.15\textwidth}p{0.07\textwidth}}
\hline
\textbf{Method} & \textbf{Social-attribution Knowledge} & \textbf{Knowledge Source} & \textbf{Instance-Specific} \\
\hline
\cite{kojima2023large} 
& \texttimes 
& Implicit (LLM prior) 
& \texttimes \\

\cite{parikh-etal-2023-exploring} 
& \texttimes 
& Human-crafted 
& \texttimes \\

\cite{v-ganesan-etal-2023-systematic} 
& \texttimes 
& External knowledge 
& \texttimes \\

\cite{10.1145/3725816} 
& Partial (context only) 
& Human-crafted 
& \texttimes \\

\textbf{Ours} 
& \textbf{Context + User Goal} 
& \textbf{LLM-generated} 
& \textbf{\checkmark} \\
\hline
\end{tabular}

\label{tab:comparison_prompting}
\end{table}

\textbf{Instruction tuning} has become an effective paradigm for using LLMs in 
diverse NLP  
tasks of AI systems in domain-specific applications, as it enables us to request the LLM with natural language prompts without any additional fine-tuning or needing additional datasets \cite{gupta2022instructdial}. 
Previous research \cite{brown2020language}
has shown that providing natural language instructions that describe the task can leverage the intrinsic knowledge of pre-trained LLMs for different NLP tasks in zero-shot settings. Recent works 
\cite{wei2022finetuned}, \cite{mishra2022crosstask}
have also demonstrated impressive performance of instruction tuning in 
general zero-shot NLP tasks. However, using this paradigm in more complex and domain-specific tasks is more challenging and requires task- and domain-specific descriptions. \textit{Chain-of-Thought (CoT)} \cite{wei2022chain} is a paradigm that is widely used to improve the performance of these models through reasoning step by step. In zero-shot setting, Zero-CoT \cite{kojima2023large} is one of the state-of-the-art methods that uses this paradigm by prompting the LLM with the statement of \textit{``Let's think step by step''}. This statement acts as a hint and directs the LLM to reason about the task step by step before the inference. However, since LLMs' reasoning processes and focus areas are unclear, social-attribution bias in LLMs can skew reasoning and undermine the Zero-CoT method's performance in social contexts. Our method mitigates this bias in the inference stage by providing prompt aids based on social-attribution knowledge (context or user's goal) of the input text 
for a given task, 
explicitly.

More recent works have focused on leveraging LLMs in more complex classification tasks. For instance, \cite{parikh-etal-2023-exploring} proposed a method for intent detection by defining different intent labels to help the LLM in inference time. \cite{v-ganesan-etal-2023-systematic} prompted the LLM with external knowledge that defines different traits to enable the LLM for zero-shot personality detection. 
Moreover, \cite{10.1145/3725816} utilized LLMs for stance detection by providing task description and context, and showed that providing the context is effective for stance detection task. Our method differs from prior work in two key aspects: (1) it explicitly models social-attribution knowledge, namely the \textit{context} and \textit{user goal} of the given text, in addition to providing task label definitions; and (2) whereas previous approaches primarily rely on human-crafted or externally defined contextual knowledge, we propose a scalable solution that leverages the LLM itself to generate this knowledge. 
For example, \cite{10.1145/3725816} 
employs a single human-crafted context shared across all instances, while our method utilizes the intrinsic knowledge of the LLM to generate instance-specific prompt aids based on social-attribution knowledge. This design enables scalability, which is essential for 
AI systems 
in practice. Table~\ref{tab:comparison_prompting} compares our method with prior context injection and external knowledge prompting approaches.

Prompt engineering is a key step in instruction tuning, and various prompt templates for different tasks have been proposed in previous works \cite{10.1145/3560815}. The LLM's faithfulness to the given context in the prompt is another perspective explored in \cite{zhou2023contextfaithful}, where the authors provide prompt strategies for this purpose. Here, we follow the guidelines of the instruction-based strategy in \cite{zhou2023contextfaithful} to use both the background knowledge of the LLM and the provided prompt aids based on social-attribution knowledge.

\section{Methodology}
In this study, we evaluate LLMs' performance in language understanding tasks requiring social context
and propose prompt strategies to reveal the models' social-attribution bias and  improve their performance in such complex tasks. Although these models performed well in simple classification tasks in prior studies \cite{wei2022finetuned}, \cite{mishra2022crosstask}, as we show, they still exhibit knowledge deficiencies in understanding the social aspects of the text in their reasoning process, leading to decreased performance. 

\begin{figure*}[t]
  \centering

\includegraphics[width=\textwidth]{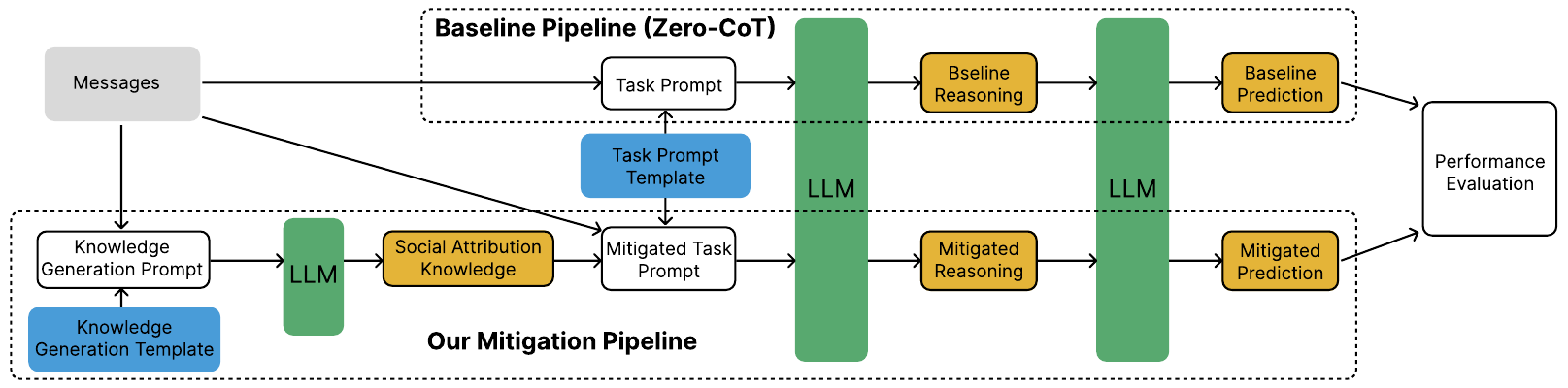}
\caption{\label{fig:our-framework}Our proposed framework to investigate the social-attribution bias of LLMs in the behavioral analytics
tasks.
}
\end{figure*}

Fig. \ref{fig:our-framework} illustrates our proposed framework to investigate the social-attribution biases of LLMs in our target tasks. In this framework, we consider the Zero-CoT pipeline, as the baseline, to first generate the task prompt based on a task prompt template, then ask the LLM to generate the baseline reasoning, and finally use the LLM to output the baseline prediction. Our mitigation pipeline adds another step before this pipeline to generate the prompt aid based on social-attribution knowledge and augment the task prompt with that. To this end, we leverage a knowledge generation template to prepare the knowledge generation prompt, and following that the LLM is used to output the social-attribution knowledge. Augmenting the task prompt with social-attribution knowledge allows us to evaluate its effect in mitigating LLM's  social-attribution bias. 
We propose two prompt strategies to help the model overcome its social knowledge gap.  
These knowledge types can be categorized into two categories as described in Table \ref{tab:prompt_strategies}. We argue that for each task, the prompt should be enriched with the aid based on suitable social-attribution knowledge. For intent detection, the model must focus on dispositional causality and pay attention to the user's goal, whereas theme detection requires incorporating domain-specific context so the model can rely on situational causality in its decision-making.

\begin{figure*}[t]
  \centering

\includegraphics[width=\textwidth]{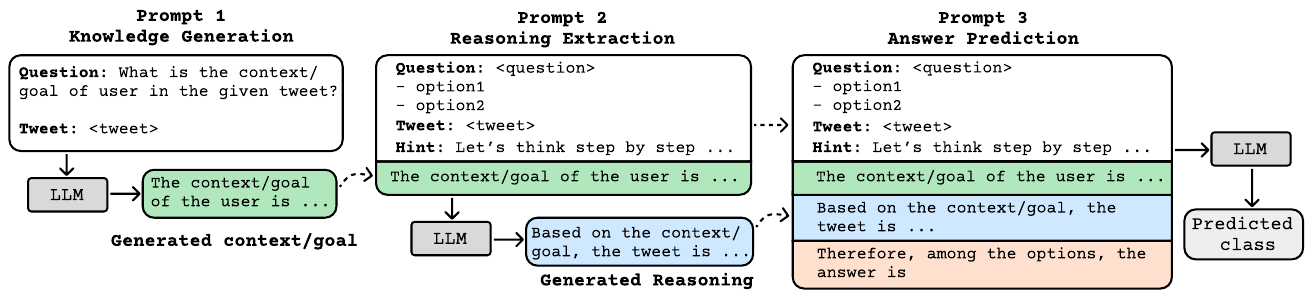}
\caption{\label{fig:architecture}Pipeline of our proposed method.
}
\end{figure*}

\begin{table}[b]
\caption{\label{tab:prompt_strategies}Motivation for designing and selecting prompt strategies based on the matching of the expected social-attribution knowledge aid for a task.}
\footnotesize
\begin{tabular}{@{}p{0.027\textwidth}p{0.065\textwidth}p{0.22\textwidth}p{0.11\textwidth}@{}}
\toprule
\textbf{Task} & \textbf{Causality} & \textbf{Description}                                       & \textbf{Prompt Strategy} \\ \midrule
\textbf{Intent}      & Dispositional                  & Model's capability of understanding the user's goal & Goal-based               \\
\textbf{Theme}       & Situational                  & Model's capability of understanding the task domain        & Context-based            \\ \bottomrule
\end{tabular}

\end{table}

\begin{table*}[t]
\caption{Samples of theme detection using Llama3 model}
\label{tab:example-theme-detection}
\footnotesize

\begin{tabular}
{@{}p{0.1\textwidth}p{0.38\textwidth}p{0.45\textwidth}@{}}
\toprule
\multicolumn{3}{p{0.9\textwidth}}{Classify the given tweet into one of the two given categories. \newline
- Donations and volunteering: Tweets requesting ... \newline
- Affected individuals: Tweets reporting injuries ... \newline
- Infrastructure and utilities: Tweets reporting ... \newline
Tweet: New to \#yyc and doing my part to clean up. Amazing support.  \#abflood \#yycflood \#calgarystrong \textless{}URL\textgreater{}}                                                                                                                                                                                                                                                                                                               \\ \midrule
              & \multicolumn{1}{|p{0.4\textwidth}|}{\textbf{Zero-CoT}}                                                                                                               & \textbf{Context-based}                                                                                                                                                                                                                                                                                              \\ \midrule
Knowledge Aid & \multicolumn{1}{|p{0.4\textwidth}|}{-}                                                                                                                      & The summary of the tweet: A newcomer to Calgary is helping with the flood cleanup efforts and is impressed with the community's support.                                                                                                                                                                \\ \midrule
Reasoning     & \multicolumn{1}{|p{0.4\textwidth}|}{Let's break down the tweet: \newline * "New to \#yyc and doing my part to clean up" - This suggests that the person is participating in cleanup efforts, which implies that \underline{they are affected} by the flood.\newline * "Amazing support" - This could be referring to the support from others, which \underline{is not directly related} to the flood.} & The tweet mentions the individual is "doing my part to clean up", which indicates \underline{they are volunteering} their time to help with the flood cleanup efforts. Additionally, the tweet expresses \underline{appreciation for the community's} \underline{support}, which suggests that the individual is part of the response efforts. \\ \midrule
Answer     & \multicolumn{1}{|p{0.4\textwidth}|}{Affected Individual \textcolor{red}{\ding{55}}} & Donations and volunteering \textcolor{green}{\ding{51}}
 \\ \bottomrule
\end{tabular}
\end{table*}

\subsection{Proposed Mitigation Pipeline}
We propose the three-step pipeline depicted in Fig. \ref{fig:architecture} to mitigate the social-attribution bias of the LLM in our target zero-shot tasks. This pipeline extends the two-step pipeline of Zero-shot CoT method \cite{kojima2023large} by adding a \textit{Knowledge Generation} step and using the generated knowledge in the \textit{Reasoning Extraction} step. First, in the \textit{Knowledge Generation} step, given the message (Tweet) the model is asked to generate the prompt aid based on the desired social-attribution knowledge (\textit{context} or \textit{goal}). Then, we formulate the generated knowledge aid as a hint in the \textit{Reasoning Extraction} step to guide the model's reasoning with this social-attribution knowledge. In this step, the model is prompted with a question and a list of options that guide the LLM to generate a reasoning for the given message based on the 
provided knowledge aid. Finally, in the \textit{Answer Prediction} step, the generated reasoning and a conclusion statement are 
appended as parts of the answer to the previous prompt, following that the LLM is asked to complete the answer by predicting the class label for the message input from
the given options.

\subsection{Prompt Strategies}
\textit{Context-based} and \textit{Goal-based} are two fundamental prompt strategies (Table~\ref{tab:prompt_strategies}) we propose to address \textit{domain} and \textit{affect}-based deficiencies for social-attribution knowledge 
respectively. 

\begin{figure*}[]
  \centering

\includegraphics[width=\textwidth]{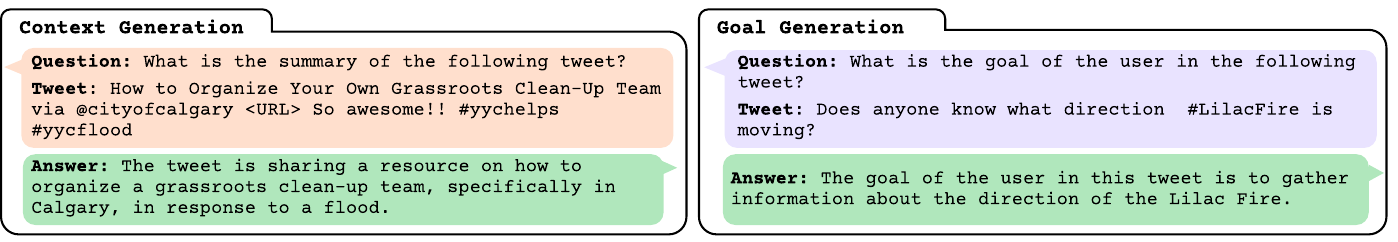}

\caption{\label{fig:generation-dialogue}Examples of Context and Goal knowledge aids generated by Llama3 model.
}

\end{figure*}

\textbf{Context-based}. In domain-specific tasks, paying attention to the context of the text for LLMs is more important than general-domain tasks. 
For instance, in the example represented in Table \ref{tab:example-theme-detection}, the tweet expresses that the user is a newcomer who is helping the cleanup effort after a flood. However, the baseline method (Zero-CoT) misinterprets the user as a resident and mentions \textit{``they are affected''}. Also, while \textit{``Amazing support''} conveys the response activities, the baseline fails to capture the causality of the situational event (community support) to the behavior of the user (appreciation), and so selects wrong label of \textit{``Affected Individual''}. 
In contrast, providing the summary of the tweet that contains contextual information such as \textit{``newcomer''} and \textit{``community's support''} in our Context-based method helps the model to correctly consider the user as a volunteer and focus on \textit{``appreciation for the community's support''} (using situational causality), classifying the tweet into \textit{``Donations and volunteering''}.
We hypothesize that explicitly providing a text’s context helps the model focus on domain-related information and rely on situational causality in reasoning. Moreover, instruction-following LLMs, thanks to training on massive corpora, effectively understand and generate the context of the given text. Our experiments show that generating a text summary effectively captures its contextual information. Therefore, we first prompt the model to generate the summary of the text (see Fig. \ref{fig:generation-dialogue}), and then put the generated summary as a hint in the reasoning extraction prompt to enrich the prompt with contextual information.

\textbf{Goal-based}. A lack of knowledge about the user's goal in writing a text can lead to misunderstandings of the text by LLMs. For instance, as we observe in Fig. \ref{fig:main-dialogue}, the baseline model fails to understand the user's goal in sharing the link, and this leads to a wrong prediction by the model.
We hypothesize that prompting the model with explicit users' goal can help the model concentrate on the main point of the given text, 
while incorporating dispositional causality to eliminate this social-attribution bias and predict the correct answer. Although instruction-following LLMs can perceive the user's goal, we argue that they may ignore that in their reasoning process. Thus, we first exploit the LLM as a goal extractor and generate the user's goal, as shown in Fig. \ref{fig:generation-dialogue}, following which we use the generated goal as the given knowledge in the prompt template.

\subsection{Prompt Templates}
\begin{figure}[tp]
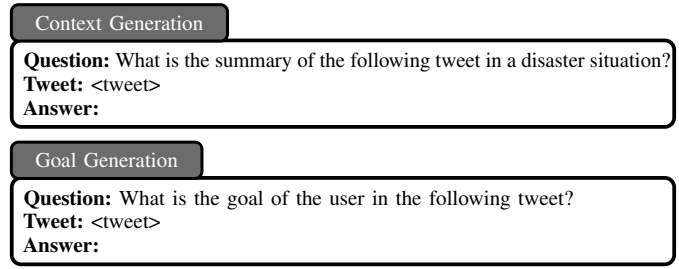

\footnotesize
  \centering
  
    \begin{prompt_box}[colback=white]{Context Generation}
    \textbf{Question:} What is the summary of the following tweet in a disaster situation?

    \textbf{Tweet:} <tweet>

    \textbf{Answer:}
    \end{prompt_box}

    \begin{prompt_box}[colback=white]{Goal Generation}
    \textbf{Question:} What is the goal of the user in the following tweet?

    \textbf{Tweet:} <tweet>

    \textbf{Answer:}
    \end{prompt_box}
  \caption{Templates for context and goal knowledge generation.}
  \label{fig:context_goal_generation_prompt}

\end{figure}

We design our templates based on multi-class classification tasks, in that the task is defined as $T = (q, t, o)$, where $q$ is a question that represents the task, $t$ is the input text for the task and $o$ is the list of labels from which the model must select the correct label.
To prompt LLMs based on the proposed strategies, we design our prompt templates for zero-shot setting. For this purpose, we first extend the task to $T = (q, t, o, H)$, in which $H$ is the hint we design for different prompt strategies. For instance, for Zero-CoT, we use the statement \textit{"Let's think step by step"} as suggested by the authors in \cite{kojima2023large}. For our proposed strategies we use a pair of $(h, G(t))$ as the hint, in that $h$ is a customized statement for the strategy derived from the statement of the Zero-CoT and $G(t)$ is a generative function of the strategy to generate the context or goal of the input $t$. Here, we use $h=$ \textit{"Let's think step by step about the user's goal"} as the statement of our \textit{goal-based} strategy and $h=$ \textit{"Let's think step by step about the summary of the tweet"} for the \textit{context-based} strategy. The designed templates for generating the context and goal knowledge aids and the template used for classification task are shown in Fig. \ref{fig:context_goal_generation_prompt} and  \ref{fig:knowledge_wise_prompt} respectively.

\section{Experiments}
To evaluate our hypotheses and research questions, we conduct experiments on two complex 
behavioral analytics tasks (intent detection and theme detection) 
and 
two settings: Mono-lingual (datasets containing English tweets) 
and multi-lingual (non-English datasets),  
during several natural disaster events.

\subsection{Datasets}
Here, we leverage various datasets in both mono-lingual (English) and multi-lingual settings to address our research questions.
For the intent detection task in mono-lingual settings, we use 
\textit{TREC IS dataset} \cite{buntain2022incident} 
and select datasets of tweets regarding two wildfire events, including \textit{Lilac} and \textit{Thomas}, to evaluate the method in a cross-event setting as well. However, to adapt the annotated dataset 
to our desired consistent labels, we select messages with \textit{GoodServices}, \textit{SearchAndRescue}, and \textit{InformationWanted} labels as the \textit{Help-seeking} samples and the messages with \textit{ServiceAvailable} label as the \textit{Help-offering} samples. For our multi-lingual experiments, we use two datasets: 1) A dataset of tweets in Haitian Creole \cite{krishnan-etal-2021-multilingual} posted during the \textit{Haiti} Earthquake, containing two classes: \textit{Requests} and \textit{Other intents}. 2) A dataset of tweets in Spanish regarding the \textit{Chile} Earthquake derived from \cite{alam2021crisisbench}. We use samples with \textit{Requests or Needs} label for binary classification, treating \textit{Not Humanitarian} as the negative class.

In the mono-lingual theme detection setting, we use the \textit{CrisisLexT26} dataset \cite{10.1145/2675133.2675242}, focusing on three themes: \textit{Donations and volunteering}, \textit{Infrastructure and utilities}, and \textit{Affected individuals}, using English tweets from two flood events, \textit{Alberta} and \textit{Colorado}. In the multi-lingual setting, we examine tweets from two earthquake events, \textit{Chile} and \textit{Italy}, in Spanish and Italian respectively \cite{alam2021crisisbench}. In both cases, the classes align with the mono-lingual datasets. Table \ref{tab:dataset-statistics} presents the dataset statistics.

\begin{table*}[]
\caption{\label{tab:dataset-statistics}Dataset Statistics}

\footnotesize
\begin{tabular}{p{0.06\textwidth}p{0.04\textwidth}p{0.11\textwidth}|p{0.68\textwidth}}
\toprule
\textbf{Task} & \textbf{Lang.} & \textbf{Event} & \textbf{Labels (\#Samples)}                       \\ \midrule
\textbf{Intent}                & Mono                  & Lilac (en)                  & Help-seeking (128), Help-offering (263)                          \\
                      &                       & Thomas (en)                & Help-seeking (156), Help-offering (177)                          \\ \cmidrule{2-4} 
                      & Multi                 & Haiti (ht)             & Requests (324), Others (196)                         \\
                      &                       & Chile (es)              & Requests or Needs (64), Others (149)                          \\ \midrule
\textbf{Theme}                 & Mono                  & Alberta (en)                & Donations and volunteering (220), Infrastructure and utilities
(199), Affected individuals (78)                         \\
                      &                       & Colorado (en)               & Donations and volunteering (149), Infrastructure and utilities
(175), Affected individuals (205)                          \\ \cmidrule{2-4} 
                      & Multi                 & Chile (es)             & Donations and volunteering (496), Infrastructure and utilities (125),
Affected individuals (377)                          \\
                      &                       & Italy (it)             & Donations and volunteering (193), Infrastructure and utilities (82),
Affected individuals (92)                          \\ \bottomrule
\end{tabular}
\end{table*}

\subsection{Experimental Setup}
In this work, we evaluate Llama3 (8B) \cite{llama3modelcard}, Mistral (7B) \cite{jiang2023mistral}, and Gemma (7B) \cite{team2024gemma} as three open-source instruction-following LLMs on our two language understanding tasks. For experiments, we set the temperature to zero for deterministic results and examine its effect in our ablation studies.

\begin{figure}[tp]
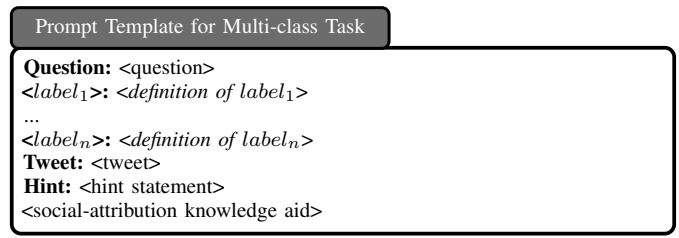

  \centering
  \footnotesize
 \begin{prompt_box}[colback=white]{Prompt Template for Multi-class Task}
   \textbf{Question:} <question>
    
    \textbf{<\textit{$label_1$}>:} <\textit{definition of $label_1$}>
    \newline
    ...
    \newline
    \textbf{<\textit{$label_n$}>:} <\textit{definition of $label_n$}>
    \newline
    \textbf{Tweet:} <tweet>
    \newline
    \textbf{Hint:} <hint statement>
    \newline
    <social-attribution knowledge aid>
    \end{prompt_box}

  \caption{Prompt template for multi-class classification task.}
  \label{fig:knowledge_wise_prompt}
\end{figure}

We use the template of Fig. \ref{fig:knowledge_wise_prompt} to generate the prompts of Zero-CoT and our proposed pipeline for classification tasks by providing the task question, class labels and their definitions, tweet, and hint statement. We additionally compare our method with Self-Debias approach \cite{gallegos-etal-2025-self} to evaluate the extent to which LLMs can mitigate bias solely through self-regulation, without relying on knowledge augmentation (our method). To implement this method, we adopt the template shown in Fig. \ref{fig:knowledge_wise_prompt} (excluding the Hint) and prompt the LLM to generate a revised conclusion using the instruction: “Remove bias from your answer by answering the question again with the category only,” following \cite{gallegos-etal-2025-self}. Further, our observations show that label order in prompts can significantly affect performance. To mitigate this bias, we evaluate all option permutations and report average results. Our ablation study highlights that our strategies are more robust to the order bias across different models and tasks.

We select two behavioral analytic tasks, intent detection and theme detection, in mono-lingual (English) and multi-lingual settings to evaluate the effectiveness of our proposed methods across tasks and languages. 
All experiments are designed as multi-class classification tasks in which the model is asked to classify the given message into given classes. Exceptionally, in the case of intent detection on the \textit{Chile} dataset which is a binary classification task, the model is asked to classify the given message into \textit{Yes} or \textit{No} classes based on the definition provided in the question. In our multi-class tasks, we define each class to help the model understand the task accurately and mitigate potential biases in its perception of class labels. Definition of class labels in different datasets of intent detection and theme detection tasks is 
provided in the supplementary materials. 
The definitions employed in our experiments are directly adopted from those provided by the dataset creators.

\subsection{Evaluation Metric}
LLMs’ bias toward social attributions and neglect of them in reasoning leads to biased responses and reduced performance. Therefore, in this work, we evaluate the bias of the methods based on the performance of the model in predicting correct labels.
Our datasets, like real-world scenarios, are not balanced in different task labels. So, in this work, we report the performance of our methods in \textit{Macro-F1} score. This eliminates the effect of imbalanced classes in the datasets.

\section{Results}
    Recall that 
    our primary objective is to assess whether 
    the intrinsic social-attribution bias of LLMs degrades their performance in our zero-shot classification tasks, and whether our proposed method reveals and mitigates this bias in their inference stage. Here, we evaluate this hypothesis on three Instruction-following LLMs in different settings to provide deeper insights into the performance of the LLM in the target tasks and answer our research questions.

\subsection{Social-attribution Bias Analysis}
To address our research questions \textit{RQ1} and \textit{RQ2}, we conducted experiments on monolingual (English) datasets to evaluate three LLMs on the target tasks and to demonstrate how our proposed prompting strategies improve baseline performance by mitigating social-attribution biases. Overall, the results in Table \ref{tab:main-results} show that \textbf{in  all target events of both tasks in the mono-lingual setting, our proposed prompt aids based on social-attribution knowledge improve the performance of baselines significantly}. This observation \textbf{confirms our \textit{RQ1} and reveals the presence of social-attribution biases in the LLMs}. For instance, in the intent detection task, while the baseline method (Zero-CoT) on the Llama3 model results in macro-F1 scores of 80.8 and 82.5 (the best results of the baseline methods across different LLMs) in the two target events, our strategy (goal-based strategy) improves them to 88.7 and 86.2 respectively. In the theme detection task, the baseline method on Mistral shows the best results (81.0 and 83.6 for Alberta and Colorado datasets respectively) whereas they are improved to 82.8 and 84.8 respectively by our strategy (context-based).
Comparing our method with Self-Debias shows that across all tasks and LLMs, our method consistently outperforms Self-Debias, with the only exception being theme detection on the Gemma model, where Self-Debias achieves competitive performance. The results also reveal that Self-Debias has an inconsistent impact across models: although it improves over Zero-CoT on Gemma, its performance is lower on Mistral. This suggests that Self-Debias is not a reliable 
debiasing approach across different LLMs on our target tasks.  

\begin{table*}[tp]
\caption{\label{tab:main-results}The performance (macro-F1 score) of our proposed strategy in enhancing the prediction accuracy of the target models in \textit{intent detection} and \textit{theme detection} tasks on mono-lingual and multi-lingual datasets. The best result is \textbf{bolded}.}
\footnotesize
\begin{tabular}
{@{}p{0.035\textwidth}p{0.03\textwidth}p{0.06\textwidth}|
 p{0.065\textwidth}p{0.075\textwidth}p{0.06\textwidth}|
 p{0.065\textwidth}p{0.075\textwidth}p{0.06\textwidth}|
 p{0.065\textwidth}p{0.075\textwidth}p{0.06\textwidth}@{}}
\toprule
\multirow{2}{*}{\textbf{Task}} &
\multirow{2}{*}{\textbf{Lang}} &
\multirow{2}{*}{\textbf{Dataset}} &
\multicolumn{3}{c|}{\textbf{Llama3}} &
\multicolumn{3}{c|}{\textbf{Mistral}} &
\multicolumn{3}{c}{\textbf{Gemma}} \\ 
\cmidrule(l){4-12}
 & & 
 & Zero-CoT & Self-Debias & Ours
 & Zero-CoT & Self-Debias & Ours
 & Zero-CoT & Self-Debias & Ours \\ 
\midrule

\underline{\textbf{Intent}} & Mono & Lilac
 & 80.8\textsubscript{$\pm$1.4} & 79.6\textsubscript{$\pm$3.2} & \textbf{88.7\textsubscript{$\pm$0.4}}
 & 77.6\textsubscript{$\pm$2.7} & 74.0\textsubscript{$\pm$3.4} & \textbf{81.8\textsubscript{$\pm$1.2}}
 & 74.0\textsubscript{$\pm$1.7} & 76.7\textsubscript{$\pm$0.4} & \textbf{83.8\textsubscript{$\pm$0.7}} \\

 & & Thomas
 & 82.5\textsubscript{$\pm$0.3} & 82.4\textsubscript{$\pm$1.1} & \textbf{86.2\textsubscript{$\pm$0.6}}
 & 79.4\textsubscript{$\pm$2.7} & 79.2\textsubscript{$\pm$2.2} & \textbf{87.4\textsubscript{$\pm$0.3}}
 & 79.3\textsubscript{$\pm$0.7} & 80.8\textsubscript{$\pm$1.1} & \textbf{86.2\textsubscript{$\pm$0.9}} \\ 
\cmidrule(l){2-12}

 & Multi & Haiti (ht)
 & 67.2\textsubscript{$\pm$0.2} & 64.1\textsubscript{$\pm$4.1} & \textbf{69.8\textsubscript{$\pm$0.5}}
 & 67.5\textsubscript{$\pm$1.7} & 61.5\textsubscript{$\pm$0.6} & \textbf{68.5\textsubscript{$\pm$0.7}}
 & 51.7\textsubscript{$\pm$1.4} & 54.0\textsubscript{$\pm$1.2} & \textbf{65.4\textsubscript{$\pm$0.8}} \\

 & & Chile (es)
 & 70.0\textsubscript{$\pm$0.4} & \textbf{72.1\textsubscript{$\pm$1.5}} & 70.9\textsubscript{$\pm$0.8}
 & 71.4\textsubscript{$\pm$0.6} & \textbf{72.4\textsubscript{$\pm$1.1}} & 71.0\textsubscript{$\pm$0.6}
 & 62.6\textsubscript{$\pm$0.5} & 62.6\textsubscript{$\pm$0.5} & \textbf{64.4\textsubscript{$\pm$0.8}} \\ 
\midrule

\underline{\textbf{Theme}} & Mono & Alberta
 & 74.0\textsubscript{$\pm$3.0} & 75.7\textsubscript{$\pm$4.9} & \textbf{79.3\textsubscript{$\pm$3.2}}
 & 81.0\textsubscript{$\pm$2.1} & 79.1\textsubscript{$\pm$3.0} & \textbf{82.8\textsubscript{$\pm$2.0}}
 & 75.5\textsubscript{$\pm$2.8} & 76.9\textsubscript{$\pm$1.7} & \textbf{77.0\textsubscript{$\pm$2.6}} \\

 & & Colorado
 & 79.4\textsubscript{$\pm$2.1} & 79.5\textsubscript{$\pm$4.2} & \textbf{83.2\textsubscript{$\pm$1.5}}
 & 83.6\textsubscript{$\pm$1.6} & 83.0\textsubscript{$\pm$2.8} & \textbf{84.8\textsubscript{$\pm$1.0}}
 & 82.6\textsubscript{$\pm$1.5} & 84.2\textsubscript{$\pm$0.7} & \textbf{84.2\textsubscript{$\pm$1.1}} \\ 
\cmidrule(l){2-12}

 & Multi & Chile (es)
 & 67.3\textsubscript{$\pm$6.2} & 65.6\textsubscript{$\pm$7.3} & \textbf{72.3\textsubscript{$\pm$2.7}}
 & 77.9\textsubscript{$\pm$2.4} & 78.2\textsubscript{$\pm$1.6} & \textbf{79.6\textsubscript{$\pm$1.8}}
 & 70.4\textsubscript{$\pm$11.8} & \textbf{78.6\textsubscript{$\pm$5.5}} & 72.0\textsubscript{$\pm$7.7} \\

 & & Italy (it)
 & 77.4\textsubscript{$\pm$6.4} & 70.3\textsubscript{$\pm$9.1} & \textbf{81.2\textsubscript{$\pm$2.5}}
 & 77.9\textsubscript{$\pm$1.7} & 76.7\textsubscript{$\pm$1.6} & \textbf{79.9\textsubscript{$\pm$1.8}}
 & 68.8\textsubscript{$\pm$10.2} & \textbf{74.5\textsubscript{$\pm$5.3}} & 72.6\textsubscript{$\pm$7.1} \\ 
\bottomrule
\end{tabular}
\end{table*}

Additionally, Fig. \ref{fig:improvement-chart} demonstrates the average of the improvements in 
the performance over 
Zero-CoT,  the baseline, by different LLMs across both target tasks and multi-lingual settings. While the results confirm the effectiveness of our strategies in both tasks, \textbf{we observe that goal-based strategy for mitigating the social-attribution bias in the intent detection task results higher margin of improvements in all  models in mono-lingual setting}. The reason behind this observation can be explored in future works from different perspectives: 1) The intensity of the LLMs' bias toward the user's goal is more than that of the LLMs' for the context of the message, so the LLMs require more knowledge about the user's goal than the context of the message. 2) The generated goal-based knowledge aid is more 
informative for the LLM to accurately understand the user intent from the user-generated message than the generated context. 

All in all, the results answer our \textit{RQ2} and \textbf{confirm that enriching the prompt with the aids based on social-attribution knowledge retrieved from the LLM alleviates the social-attribution bias of the model to improve its classification performance in our target tasks.} The results highlight the importance of paying attention to the social-attribution biases in developing LLM-based AI systems for a domain.  
Additionally, our proposed method shows potential for mitigating them,  
especially in 
English 
content setting. 
 
\subsection{Multi-lingual  Analysis}
\begin{figure*}[!t]
\centering
\subfloat[]{\includegraphics[width=0.35\textwidth]{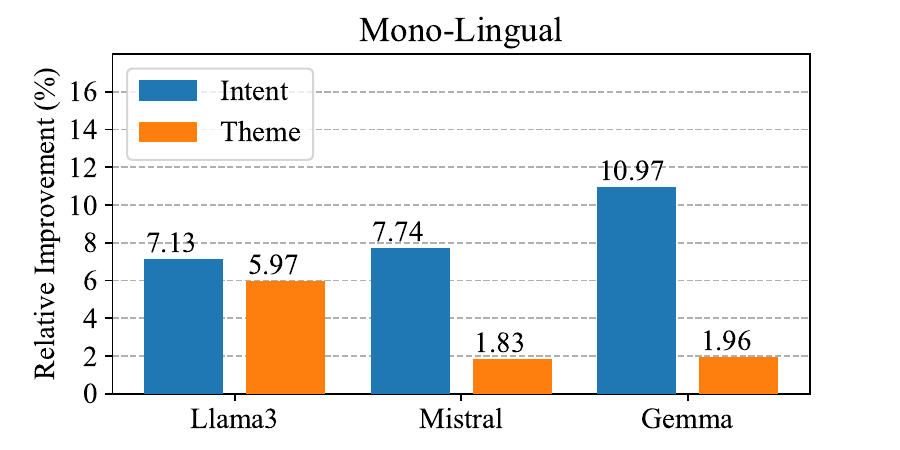}%
\label{fig:improvement-mono-lingual}}
\hfil
\subfloat[]{\includegraphics[width=0.35\textwidth]{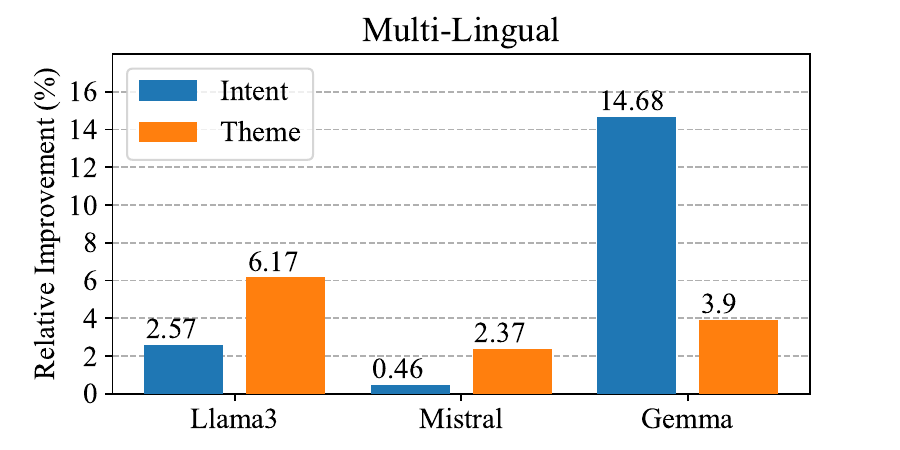}%
\label{fig:improvement-multi-lingual}}
\caption{The average of relative improvement of our strategies over Zero-CoT by different LLMs across the target tasks.}
\label{fig:improvement-chart}
\end{figure*}

We examined our proposed strategies on several multi-lingual datasets to address our \textit{RQ3}. The results of Table \ref{tab:main-results} \textbf{show the effectiveness of our method across different tasks, models, and languages in most cases (8 out of 12), consistently outperforming Zero-CoT and Self-Debias.} While Self-Debias occasionally surpasses our method for specific model-language combinations, its gains are marginal and  inconsistent. Overall, our approach provides more reliable and generalizable improvements across tasks, models, and languages, effectively mitigating social-attribution bias.
Comparing the results of the improvement by our method over Zero-CoT, the 
baseline, 
in Fig. \ref{fig:improvement-chart}, (a) and (b) indicate 
that \textbf{while our strategies are effective in multi-lingual settings as well, 
they demonstrate 
greater consistency in the English datasets.} The main reason for this observation would be that different LLMs, trained on varying corpora, have distinct abilities in understanding different languages, particularly low-resource languages like Haitian Creole. For instance, in comparison to Chile dataset (Spanish language), we observe greater margins of improvement over the baseline by our method for intent detection on Haiti dataset (Haitian Creole language) across all LLMs, specifically on Gemma by 26.5\%. Another direction that can be explored in future works is how social factors of a language can be captured and transferred across languages. In our case, when we generate the user's goal from a non-English message, we cannot guarantee that the LLM captures and reflects all social-attribution knowledge of the original language in the generated prompt aids in English.
The rest of the analyses 
use 
Zero-CoT as the primary baseline, as it underlies our method and shows more consistent performance in the monolingual setting.

\subsection{Comparative Analysis for  LLMs} 

\begin{figure*}[!t]
\centering
\subfloat[]{\includegraphics[width=0.35\textwidth]{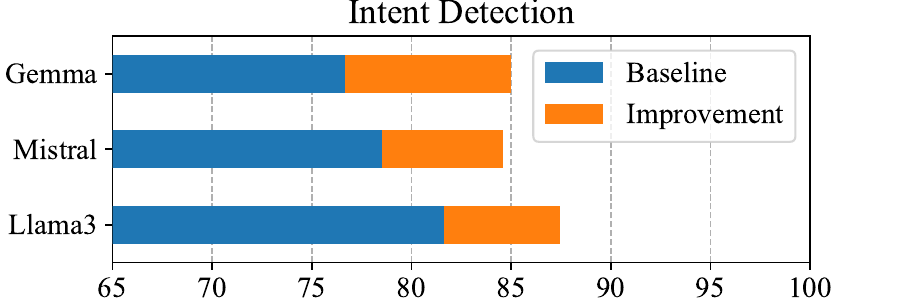}%
\label{fig:improvement-llms-intent}}
\hfil
\subfloat[]{\includegraphics[width=0.35\textwidth]{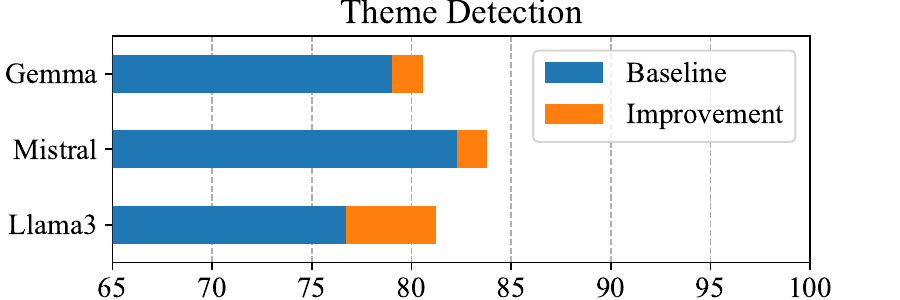}%
\label{fig:improvement-llms-theme}}
\caption{The average improvement of our strategies over Zero-CoT by different LLMs in mono-lingual setting.}
\label{fig:improvement-llms}
\end{figure*}

Different LLMs, due to their differences in the architecture and training stage, are expected to have different capabilities in understanding of the task and input. For instance, based on the result in Table \ref{tab:main-results}, while Llama3 shows the best result for Zero-CoT in the intent detection task on the mono-lingual datasets, Mistral outperforms the other models in the theme detection task. Fig. \ref{fig:improvement-llms} compares the average of improvements (across two datasets) by our proposed method over Zero-CoT using different LLMs on the target tasks in the mono-lingual setting. The result demonstrates that \textbf{the LLMs have different capabilities in different tasks, and none of them is predominant in all tasks}. Moreover, we observe that \textbf{the LLMs leverage the provided prompt aids based on social-attribution knowledge differently}. For instance, while Mistral shows better performance than Gemma in the baseline for the intent detection task, the latter performs better in the presence of our goal-based knowledge aids. We address two potential reasons behind this observation for further exploration in future works: 1) Data distribution may play a significant role, as observed in Table \ref{tab:main-results}. For the Lilac dataset, our method achieves greater improvement in the baseline performance with Gemma compared to Mistral, whereas the Thomas dataset shows the opposite trend. 2) The way models attend to the context of input text could also be influential. However, exploring this aspect is challenging due to the black-box nature of these models. In summary, there is no universally superior LLM for all tasks, so engineers need to select the proper LLM for their target tasks in designing computational social systems.

\subsection{Robustness Analysis: Temperature Parameter}
\begin{figure*}[!t]
\centering
\subfloat[]{\includegraphics[width=0.25\textwidth]{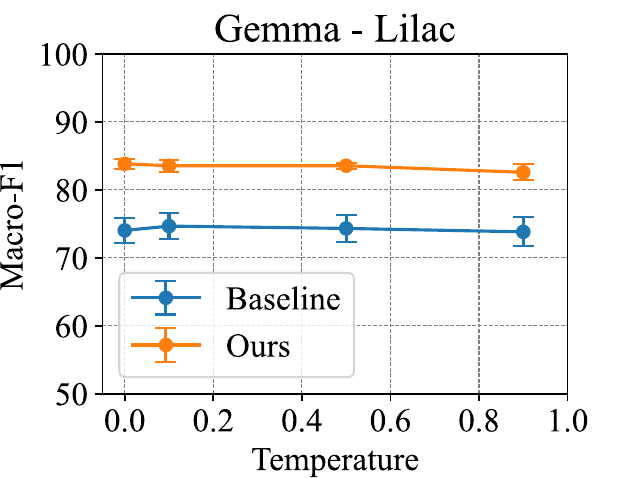}%
\label{fig:temperature-analysis-gemma-lilac}}
\hfil
\subfloat[]{\includegraphics[width=0.25\textwidth]{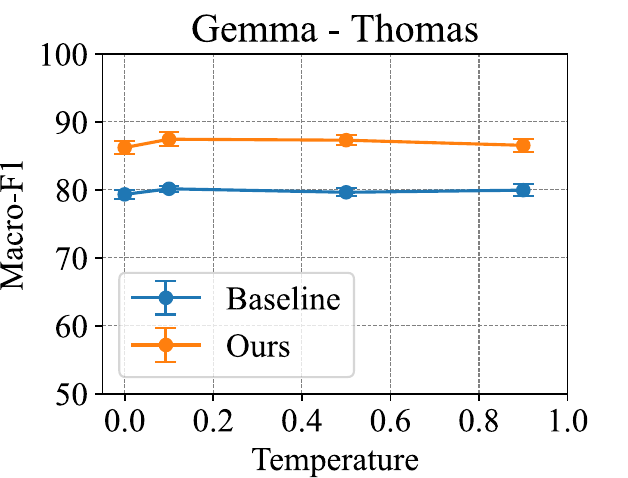}%
\label{fig:temperature-analysis-gemma-thomas}}
\subfloat[]{\includegraphics[width=0.25\textwidth]{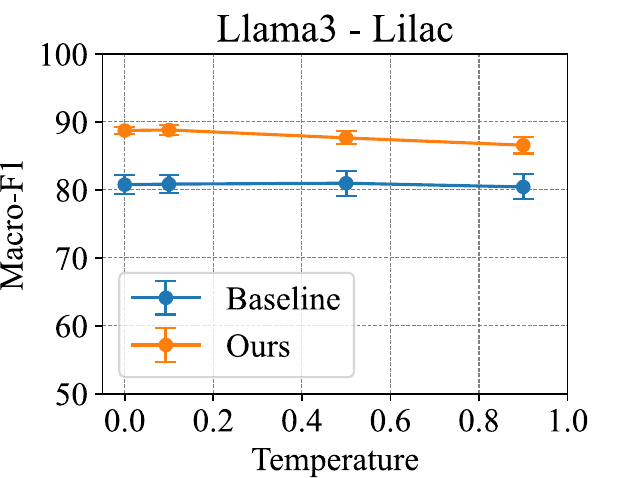}%
\label{fig:temperature-analysis-llama3-lilac}}
\hfil
\subfloat[]{\includegraphics[width=0.25\textwidth]{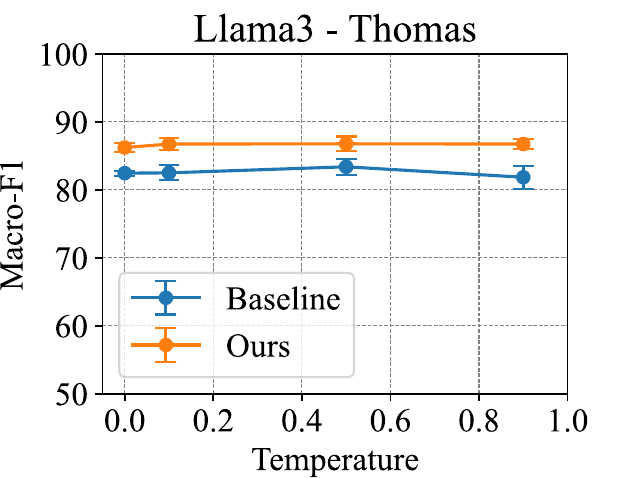}%
\label{fig:temperature-analysis-llama3-thomas}}
\caption{The effect of temperature parameter on the model performance in the intent detection task.}
\label{fig:temperature-analysis}
\end{figure*}

    Higher temperature values allow LLMs to generate more creative responses, potentially affecting performance. We evaluate this parameter's impact on the results of the intent detection task across two models, Llama3 and Gemma, by setting temperature to 0.1, 0.5, and 0.9, and comparing results with the deterministic setting (temperature = 0). We ran each experiment five times and report the average performance and standard deviation for both models on the Lilac and Thomas datasets (Fig. \ref{fig:temperature-analysis}).
    Overall, although minor fluctuations in model performance are observed at different temperatures (e.g., Fig. \ref{fig:temperature-analysis-gemma-thomas}, \ref{fig:temperature-analysis-llama3-lilac}), changing the temperature has an insignificant impact across models and datasets, with both methods (Baseline and Ours) showing the same trend. \textbf{This confirms the robustness of our observations across different temperature values}. Additionally, we observe that the standard deviation of our method is lower than the baseline in all experiments of this task (shorter candles in Fig. \ref{fig:temperature-analysis}). \textbf{This observation reveals that our proposed prompt strategies are less prone to the biases that the temperature parameter adds to the inference stage of the \textit{Zero-CoT} method}. Thus, our method can empower engineers to develop more robust solutions.

\subsection{Robustness Analysis: Order Bias}

\begin{figure}[t]
  \centering 
\includegraphics[width=0.5\textwidth]{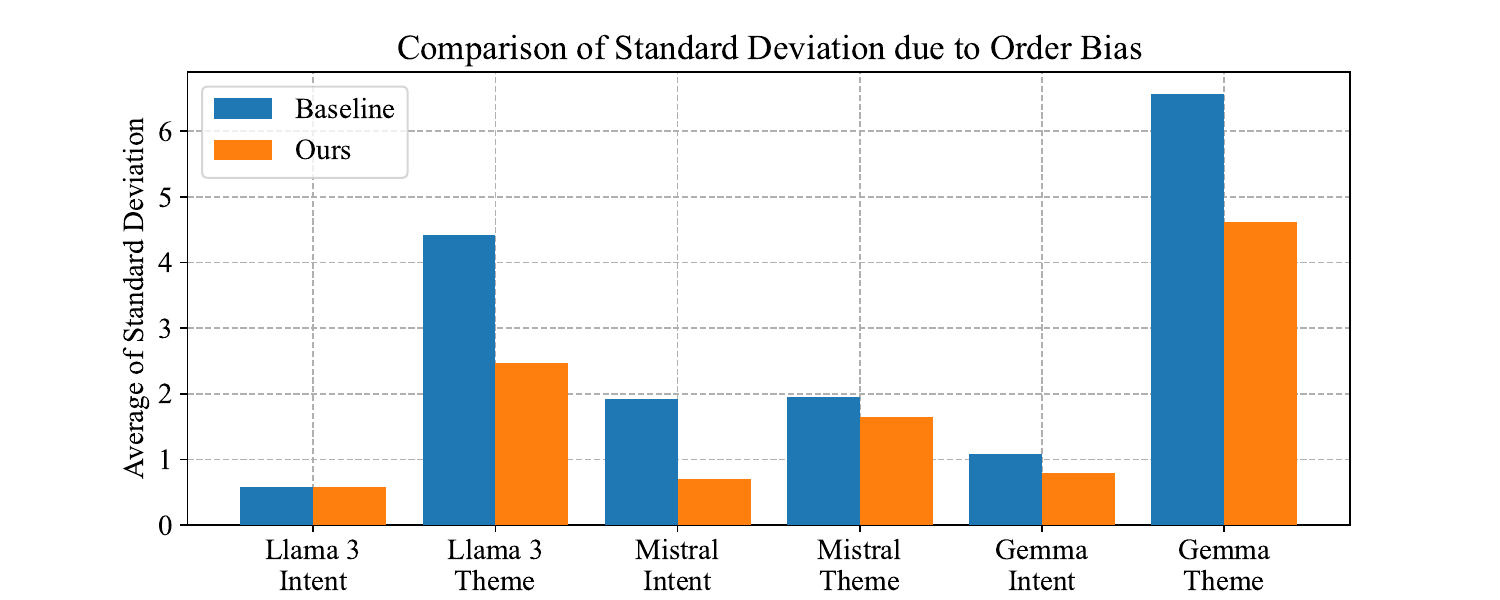}
\caption{\label{fig:order-bias}Comparison of the order bias of methods based on the standard deviation of Macro-F1 scores across model/tasks.
}
\end{figure}

The order of the given options 
may induce
a bias in the 
model's  
reasoning and performance.  
To address this issue, we report the average performance of 
a model (Macro-F1) across all combinations of the options given. 
We evaluate the robustness of our strategies toward order bias based on the standard deviation of model performance across different order settings. 
Fig. \ref{fig:order-bias} 
shows
the average of the standard deviation of the models' performance (on all datasets) for different model/ task settings. We observe that in all cases our strategies result in lower standard deviation, 
confirming 
the robustness of our method 
compared to Zero-CoT (baseline). 

\subsection{Cross-Model Analysis}
Table ~\ref{tab:self-cross-results} shows the results of our experiments to evaluate the cross-model impact of generating knowledge with another model on the performance of our method. Across both tasks and models, the performance differences between the Self and Cross settings are generally small, indicating that our method does not strongly depend on whether the knowledge is generated by the same model or by another model. While minor variations are observed—such as slight advantages for Self in intent detection and for Cross in some theme detection cases—the overall trends remain consistent. \textbf{These results suggest that our approach is robust to the source of generated knowledge, supporting its generality and applicability in cross-model settings.}

\begin{table}[tp]
\caption{\label{tab:self-cross-results}The performance (macro-F1) comparison between the impact of using knowledge generated by model itself (Self) and by another model (Cross) for Llama3 and Gemma.}
\footnotesize
\begin{tabular}
{@{}p{0.035\textwidth}p{0.055\textwidth}|
 p{0.065\textwidth}p{0.075\textwidth}|
 p{0.065\textwidth}p{0.075\textwidth}@{}}
\toprule
\multirow{2}{*}{\textbf{Task}} &
\multirow{2}{*}{\textbf{Dataset}} &
\multicolumn{2}{c|}{\textbf{Llama3}} &
\multicolumn{2}{c}{\textbf{Gemma}} \\ 
\cmidrule(l){3-6}
 & &
 Self & Cross &
 Self & Cross \\ 
\midrule

\underline{\textbf{Intent}} & Lilac
 & \textbf{88.7\textsubscript{$\pm$0.4}} & 84.5\textsubscript{$\pm$0.8}
 & 83.8\textsubscript{$\pm$0.7} & \textbf{86.0\textsubscript{$\pm$0.7}} \\

 & Thomas
 & \textbf{86.2\textsubscript{$\pm$0.6}} & 85.0\textsubscript{$\pm$0.7}
 & \textbf{86.2\textsubscript{$\pm$0.9}} & 85.6\textsubscript{$\pm$0.2} \\ 
\midrule

\underline{\textbf{Theme}} & Alberta
 & 79.3\textsubscript{$\pm$3.2} & \textbf{80.3\textsubscript{$\pm$1.8}}
 & \textbf{77.0\textsubscript{$\pm$2.6}} & 76.8\textsubscript{$\pm$2.1} \\

 & Colorado
 & 83.2\textsubscript{$\pm$1.5} & \textbf{83.7\textsubscript{$\pm$1.5}}
 & \textbf{84.2\textsubscript{$\pm$1.1}} & 83.8\textsubscript{$\pm$1.3} \\

\bottomrule
\end{tabular}
\end{table}

\subsection{Human Evaluation of AI-generated Knowledge}
To further understand the effectiveness of our approach, we conduct a human evaluation to assess the quality of the AI-generated knowledge used in our method. For each task, we select 60 samples per model (180 in total), stratified by labels and datasets, to ensure balanced coverage. We evaluate the generated knowledge using two criteria: Faithfulness, which measures whether the knowledge is factually consistent with and supported by the given tweet, and Helpfulness, which assesses whether the knowledge provides useful information for correctly answering the task. Table \ref{tab:human-eval} shows that the generated knowledge is highly faithful and helpful across both tasks and models. Faithfulness remains consistently high (0.98), indicating strong factual alignment with the content of input messages and minimal hallucination. Helpfulness scores are also high, with minor variations across models and tasks, suggesting that the knowledge effectively supports task-specific reasoning. Overall, these results validate the quality and reliability of the generated knowledge underlying our method.

\begin{table}[tp]
\centering
\caption{\label{tab:human-eval}Human evaluation results of AI-generated knowledge quality across tasks and models.}
\footnotesize
\begin{tabular}
{@{}p{0.07\textwidth}p{0.07\textwidth}|
 p{0.13\textwidth}p{0.13\textwidth}@{}}
\toprule
\textbf{Task} & \textbf{Model} & \textbf{Helpfulness (\%)} & \textbf{Faithfulness (\%)} \\
\midrule
\underline{\textbf{Intent}} & Llama3  & 0.88 & 0.98 \\
                           & Mistral & 0.88 & 0.98 \\
                           & Gemma   & 0.85 & 0.98 \\
\midrule
\underline{\textbf{Theme}}  & Llama3  & 0.85 & 0.98 \\
                           & Mistral & 0.95 & 0.98 \\
                           & Gemma   & 0.88 & 0.98 \\
\bottomrule
\end{tabular}
\end{table}

\section{Conclusion}
In this paper, we explored the presence of social-attribution biases toward two social factors, including message context and user goal, in three state-of-the-art LLMs on zero-shot domain-specific classification tasks. We proposed novel prompt strategies to reveal these biases and leverage the intrinsic knowledge of LLMs to mitigate their social-attribution biases in the inference time. Our work focused on two strategies including goal-based and context-based strategies for the intent detection and theme detection tasks respectively. 
Our experiments conducted in mono-lingual (English) and multi-lingual settings of the disaster domain showed that our strategies outperformed the  
baseline across all tasks on English datasets and most multi-lingual datasets. 
Our results revealed the presence of social-attribution biases in the  
open-source instruction-following LLMs-- Llama3, Mistral, and Gemma, 
highlighting  
the need to consider 
social-attribution biases when LLM-based AI systems are designed for social computing. Also, while the results confirmed the effectiveness of our method in improving the performance, 
we observed that there is no single LLM that works for all tasks. Although the goal-based strategy showed a higher margin of improvement in intent detection on all models compared to the context-based strategy in the theme detection task, further exploration is needed to understand how different LLMs benefit from different prompt aids in mitigating social-attribution bias. Besides, our method demonstrated more robust results than the baseline across various temperature values and option orderings, 
providing a suitable solution to design AI-powered computational social systems.

\section{Limitations and Future Work}
We address the following limitations of our study to provide insights for future research:
1) This work did not focus on providing the optimal prompt templates for different tasks, but rather 
explored the effectiveness of the proposed strategies on the manually crafted templates to reveal the existence of social-attribution biases in the LLMs and mitigate them.  
Future studies can explore  
automatic prompt template  generation. 
2) We focused on 
two social aspects of the message (\textit{goal} and \textit{context}) to highlight and mitigate the social-attribution bias of LLMs in the intent and theme detection tasks of the disaster domain. Examining the impact of additional social attributes and their combinations as well as exploring more effective mitigation strategies on various tasks could be key future work directions.
3) Due to the uneven nature of real-world disaster events, some datasets used in this work are small and imbalanced, which can lead to higher variability in model performance. 
4) Since the order of 
answer options in the prompt can bias LLM's reasoning in the result, we mitigate its effect by averaging the result over all combinations of the options. However, this approach may not extend to a large number of options. Therefore, future works can 
explore the solutions for mitigating this type of bias as well. 5) Our method requires additional LLM calls to generate task-specific social-attribution knowledge for each instance, introducing some latency and computational overhead. However, this overhead is linear and future work can 
mitigate it
via batching or parallel processing.

\bibliographystyle{IEEEtran}
\bibliography{refs}

\end{document}